\documentclass[runningheads]{llncs}

\usepackage{graphicx}
\usepackage{cite}
\usepackage{dblfloatfix}
\usepackage{todo}
\usepackage{xcolor}
\usepackage{subfig}
\usepackage{bm}
\usepackage{cleveref}

\newcommand*\samethanks[1][\value{footnote}]{\footnotemark[#1]}

\begin{document}

\title{Identity Recognition in Intelligent Cars with Behavioral Data and LSTM-ResNet Classifier \thanks{This work was not supported by any organization}}
\titlerunning{Identity Recognition with Behavioral Data}
%
\author{Michael Hammann\thanks{Both authors contributed equally to this work}\inst{1} \orcidID{0000-0001-7776-850X} \and
	Maximilian Kraus\samethanks\inst{2} \orcidID{0000-0003-1395-1780} \and
	Sina Shafaei\inst{3}\orcidID{0000-00002-9381-0197} \and
	Alois Knoll\inst{4}}

\authorrunning{Hammann et al.}
%
\institute{Department of Informatics, Technical University of Munich, Germany 		 \email{michael.hammann@tum.de}\\ \and
	Department of Informatics, Technical University of Munich, Germany
	\email{maximilian.kraus@tum.de}\\
	 \and Department of Informatics, Technical University of Munich, Germany \and
	Department of Informatics, Technical University of Munich, Germany}

\maketitle              
\vspace{-2ex}
\begin{abstract}
	
Identity recognition in a car cabin is a critical task nowadays and offers a great field of applications ranging from personalizing intelligent cars to suit drivers' physical and behavioral needs to increasing safety and security. However, the performance and applicability of published approaches are still not suitable for use in series cars and need to be improved. In this paper, we investigate Human Identity Recognition in a car cabin with Time Series Classification (TSC) and deep neural networks. We use gas and brake pedal pressure as input to our models. This data is easily collectable during driving in everyday situations. Since our classifiers have very little memory requirements and do not require any input data preproccesing, we were able to train on one \textit{Intel i5-3210M} processor only. Our classification approach is based on a combination of LSTM and ResNet. The network trained on a subset of NUDrive outperforms the ResNet and LSTM models trained solely by 35.9 \% and 53.85 \% accuracy respectively. We reach a final accuracy of 79.49 \% on a 10-drivers subset of NUDrive and 96.90 \% on a 5-drivers subset of UTDrive.

\end{abstract}

\section{Introduction} \label{sec:intro}

Identity recognition in a car cabin is an important task in the automobile industry nowadays. Done accurately, it can provide drivers and passengers with a better driving experience, increased safety and security and create a personalized driving environment. This personalization ranges from adjusting seat positions for increased comfort to individual temperature and music preferences in order to create the best driving experience possible \cite{garzon2011personal, baltrunas2011incarmusic}. Nowadays these settings are manually handled by the driver itself. Identity recognition in cars is one way to deal with these tasks automatically, resulting in the minimization of driver distractions \cite{donmez2003taxonomy}. Also, identity recognition increases security of vehicles by requiring driver authorization which reduces the risk of theft and misuse of vehicles \cite{pingat2013real}.

Above all, identity recognition makes completely new in-car tasks possible. Secure mobile transactions, such as mobile banking, cryptocurrency transactions or secured email traffic, are examples in this new area of application, especially in shared cars \cite{erdougan2005multi}. But not only drivers and passengers benefit from such a solution. Insurance companies can use this information to individually calculate the insurance premium based on a driver's behavior and adjust it accordingly \cite{jarvis2015insurance}.

Driver identification is a special form of human identity recognition and can be accomplished in several ways. The identification problem can be tackled by using Time Series Classification (TSC), basically classifying data points over time based on their behavior. In the following, we mainly focus on behavioral feature based recognition and their applicability in a car cabin. Our behavioral features are specific characteristics which describe an individual driving style over time. 

In this paper, we introduce a novel TSC approach for identity recognition in a car cabin. We train a ResNet as well as a Long Short-Term Memory (LSTM) network solely on our data and show that combining both networks in one classifier increases accuracy significantly while maintaining a suitable interference time. In contrast to previous work using one or multiple Gaussian Mixture Models (GMMs) to model each drivers' behavior, we only need to train one network which is able to classify all drivers. Our approach is suitable for an embedded in-car solution.  

\section{Related Work} \label{sec:related_work}

\subsection{Identity Recogntion Based on Facial Features}
One of the earliest approaches in human face recognition by computers was made by Brunelli and Poggio (1993) \cite{brunelli1993face}. They compare two different algorithms: Using geometrical facial features and a template matching method. For the geometrical approach 35 features are automatically extracted, recognition is then performed by a Bayes classifier. The template matching algorithm uses different grey-scaled facial region instances and compares them with the already known database images in order to get matching scores. These methods are the base for many other approaches \cite{lao20003d, ballihi2012boosting}.

In recent years, deep learning has become more and more popular for face recognition tasks \cite{sun2013hybrid, sun2014deep, taigman2014deepface}. With increasing computing power, deeper neural networks and especially Convolutional Neural Networks (CNNs) like GoogleNet \cite{szegedy2015going}, VGGNet \cite{simonyan2014very} and ResNet \cite{he2016deep} have led to major performance gains on general object recognition, image segmentation and classification. CNNs take advantage of Batch Normalization (BN) \cite{ioffe2015batch} and Dropout layers \cite{srivastava2014dropout}.  Sun et al. (2015) investigated the use of VGGNet and GoogleNet for face recognition tasks \cite{sun2015deepid3}. This approach called DeepID3 reaches 99.53 \% face verification accuracy and 96.0 \% closed set face identification accuracy with the LFW dataset \cite{huang2008labeled}.

The greatest concern with such approaches is that today's iris and face recognition algorithms can be tricked by spoofing attacks, such as photographs, videos or mask attacks \cite{kose2013vulnerability, hadid2014face, akhtar2011face}. A secure algorithm must be able to distinguish between real faces and spoofed faces. Since security is one of the core arguments of our motivation, methods based on facial recognition are not suitable.
\subsection{Identity Recognition Based on Behavioral Features}
Using behavioral features is another possibility to achieve driver identification. GMMs using pedal pressure signals are the most common method to model the behavior of individual drivers \cite{nishiwaki2007driver, benli2008driver}. Nishiwaki et al. (2007) followed a similar approach and achieved an identification rate of 76.8 \% on a dataset of 276 drivers by transforming the data into the cepstral domain. Benli et al. (2008) modeled multiple GMMs per driver for different modalities to increase confidence of the classification and examined different data fusion techniques \cite{benli2008driver}. Trainable combiners achieved in general a much better performance than fixed decision techniques. K-Nearest-Neighbors shows errors as low as 0.35 \% on a 100 person subset of the CIAIR database\cite{kawaguchi2005ciair}.

Nevertheless, GMMs come with one key disadvantage for our area of interest: Performing training of many GMMs becomes computationally expensive. The technique relies on an external server and is therefore not suitable for an embedded in-car solution \cite{martinez2015driving}.

Martinez et al. (2015) propose to use a single-hidden-layer feedforward neural network for driver identification \cite{martinez2015driving}. For the whole dataset of 11 drivers an accuracy of 76 \% was reached. There work follows a similar approach to ours - but still, simple feedforward networks do not make use of time series correlations, which are an important constraint for high level recognition rates in such cases.

\subsection{Time Series Classification with Deep Learning}
Wang et al. (2017) extended a ResNet model for TSC \cite{wang2017time}. This model performs classification on raw data and does not need preprocessing or handcrafting features. It combines a fully convolutional net with ResNet and shows competitive performance compared to state-of-the-art tools like the bag-of-features framework(TSBF) or the Bag-of-SFA-Symbols (BOSS) \cite{baydogan2013bag, schafer2015boss}.

\section{Proposed Method and Network Architecture} \label{sec:proposed_method}
Our approach is based on TSC with deep neural networks since they tend to have better generalization capabilities than classical methods. Deep classifiers already showed success in fields such as speech or video recognition \cite{abdel2014convolutional, donahue2015long}. Such classifiers will also perform well in our area of application since audio and video data share similar time dependent structures with time series data \cite{fawaz2019deep}. Furthermore, CNNs can easily handle multivariate time series and the datasets can be comparatively small since patterns only have to be detected in one dimension instead of the common two dimensions for pictures \cite{fawaz2019deep}. We use a window-based method since we want to be able to identify drivers as fast as possible. 

We have built three different classifiers which we will introduce in the following.

\subsection{LSTM networks}\label{sec:LSTM}
Long Short-Term Memories (LSTM) optimize existing recurrent neural networks by adding a memory cell in order to learn long-term temporal dependencies \cite{hochreiter1997long}. Consequently, LSTMs can deal well with sequential data and can be used for speech recognition, video classification or music composition among other things \cite{graves2013speech, yue2015beyond, choi2016text}.
Our LSTM classifier consists of two layers of stacked LSTMs and a Dropout layer in between. We use a many to one architecture, where only the final output of the LSTM is used as input to a Dropout layer before the final linear layer, which maps to the desired output size. The size of the hidden dimension is 75.

\subsection{ResNet Classifier}
We took the architecture of the work of Wang et al. (2017) \cite{wang2017time} as baseline and changed the structure and size to fit to our problem.

\begin{figure*}[thpb]
  \centering
  \includegraphics[width=\linewidth]{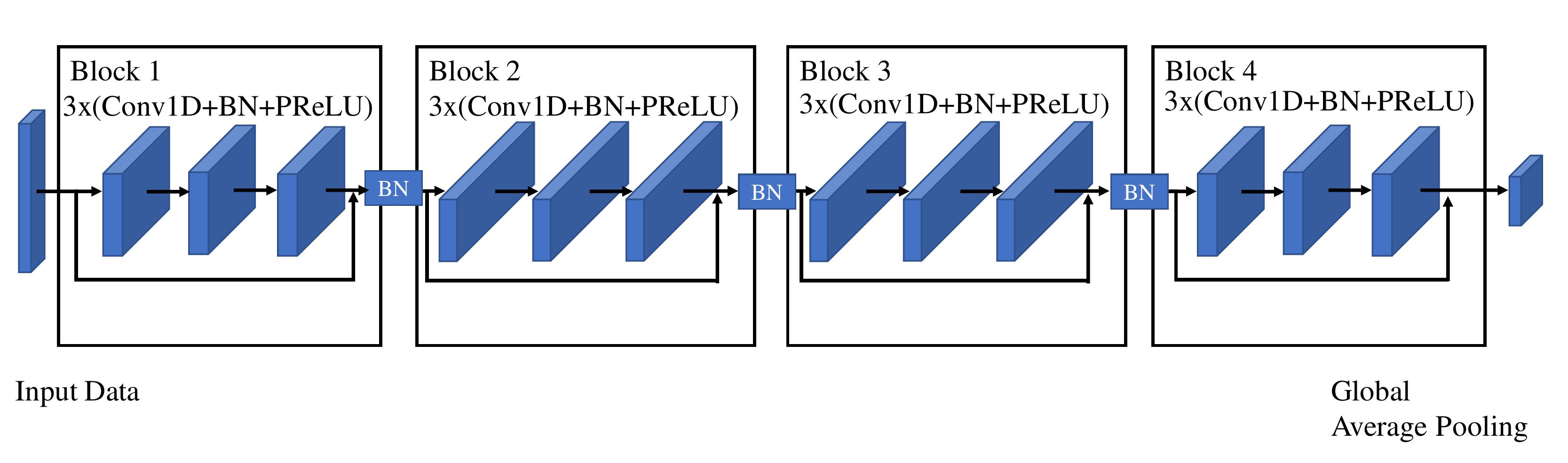}
  \caption{ResNet Architecture}
  \label{fig:RESNET}
\end{figure*}

In \Cref{fig:RESNET} our basic architecture can be seen. We use BN and PReLU activation functions instead of the ReLU functions used in \cite{wang2017time}. PReLU generalizes ReLU by learning the shape of the activation function and improves model fitting with few extra computational cost \cite{he2015delving}. A basic block is given by \Cref{eq:basic_block}.

\begin{equation} \label{eq:basic_block}
\begin{array}{l}
y = \textbf{W} \ast \textbf{x} + \textbf{b}\\ 
s = BN(y)\\
h = PReLU(s)
\end{array}
\end{equation}

A residual block is given by \Cref{eq:residual_block}. Let \(Block_k(o)\) be a convolutional block with the number of features maps \(k\).
\begin{equation} \label{eq:residual_block}
\begin{array}{l}
h_1 = Block_{k_1}(x)\\
h_2 = Block_{k_2}(h_1)\\
h_3 = Block_{k_3}(h_2)\\
y = h_3 + x\\
\hat{h} = PReLU(y)
\end{array}
\end{equation}

We use a kernel size of \{7,5,3\}, \{128,256,256,128\} feature maps, a stride of 1 and zero-padding for each residual block respectively. Furthermore, we drop the PReLU activation at the end of the last residual block. This architecture is followed by a global average pooling layer \cite{lin2013network} and a softmax layer for classification. Similar to the work of Li et al. (2018) \cite{li2018understanding}, we place a Dropout layer with \(p=0.2\) directly in front of the softmax layer to reduce variance shift.

\subsection{LSTM-ResNet Classifier}
\begin{figure*}[thpb]
	\centering
	\includegraphics[width=\linewidth]{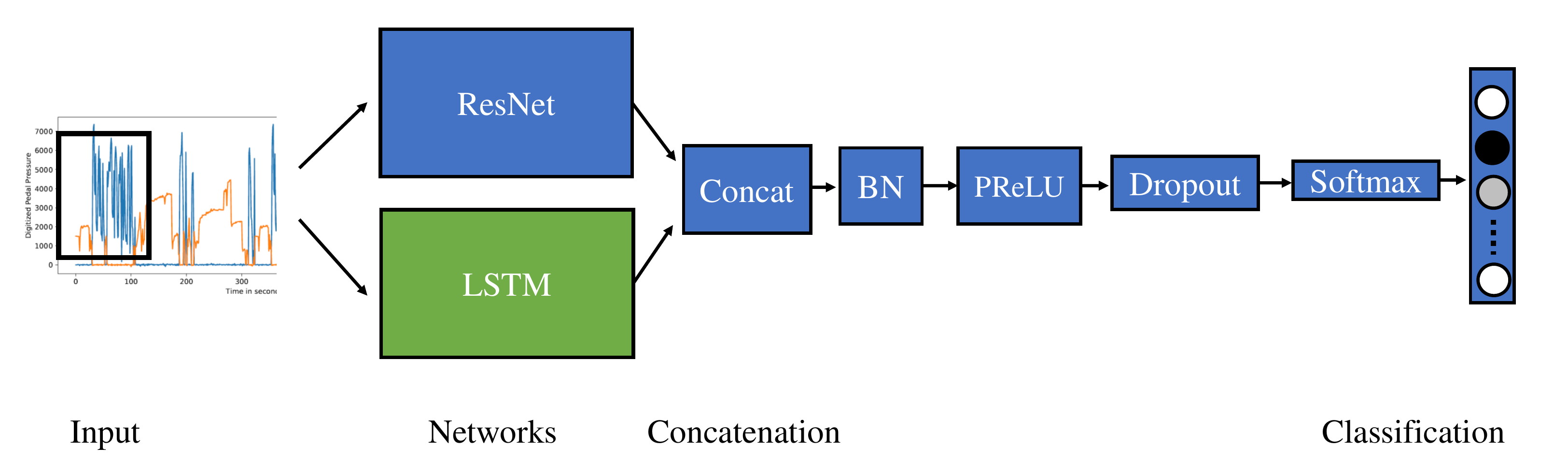}
	\caption{LSTM-ResNet Architecture}
	\label{fig:LSTM-RESNET}
	\vspace{-8.00mm}
\end{figure*}

\vspace{-1.00mm} 
LSTM and ResNet work differently with time series data: ResNet uses convolutions which can be seen as applying and sliding filters over the time series \cite{fawaz2019deep}. The result of all filters is a time series with as many dimensions as filters used. This process results in more discriminative features which are useful for classification. LSTMs are able to learn and remember over sequences of input data and do not need convolutions to aquire discriminative features. They are able to learn "long-term dependencies" which are important for time series classification. To take advantage of both network architectures, we decided to combine the LSTM and ResNet models. We remove the softmax layer as well as the Dropout layer in both classifiers. The features obtained by each of both classifiers are concatenated and fed into a convolutional 1D layer with output size 64. We use BN, PReLU and a final Dropout layer with \(p=0.2\) afterwards. This is followed by a softmax layer for classification. The full architecture can be seen in \Cref{fig:LSTM-RESNET}.

\section{Experiments} \label{sec:Experiments}
\subsection{Datasets}
We train on subsets of the UTDrive \cite{hansen2017driver,angkititrakul2009utdrive} and NUDrive \cite{takeda2011international} datasets. Each dataset contains multiple drives of each driver for five and ten drivers respectively. 

The UTDrive data was provided at a sampling rate of 100 Hz. The NUDrive data was collected at a sampling rate of 1 kHz and was then digitized into a signed 16-bit Little Endian format. The pedal data is collected as pedal pressure (kgf: kilogram force), mapped to 0-5 V and linearly digitized in the range 0-32767. The driving signals are mapped as shown in \Cref{tab:NUDrive}.

\begin{table}
	\vspace{-5.00mm}
	\caption{NUDrive data formats}
	\begin{center}
		\begin{tabular}{|c|c|c|c|}
			
			\hline
			Signal & Range & Mapped to & Digitized range \\
			\hline
			Brake pedal pressure & 0-50 kgf & 0-5 V & 0-32767 \\
			Accelerator pedal pressure & 0-30 kgf & 0-5 V & 0-32767 \\
			\hline
			
		\end{tabular}

	\end{center}

	\label{tab:NUDrive}
	\vspace{-5.00mm} 

\end{table}

\begin{figure}
	\vspace{-8.00mm} 
	\centering
	\begin{tabular}{@{}c@{}}
		\includegraphics[width=\linewidth]{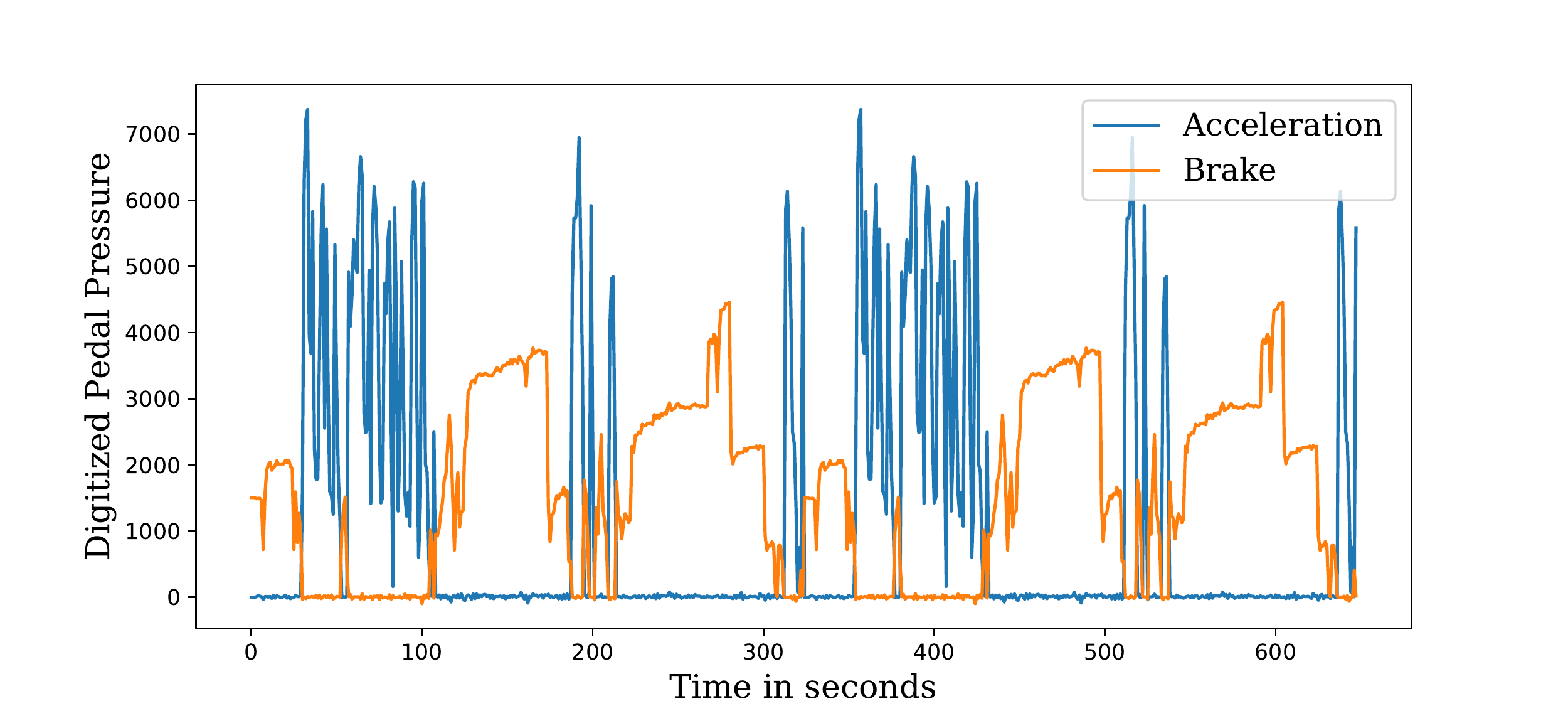} \\[\abovecaptionskip]
		(a) Driver 1 
	\end{tabular}

	\begin{tabular}{@{}c@{}}
		\includegraphics[width=\linewidth]{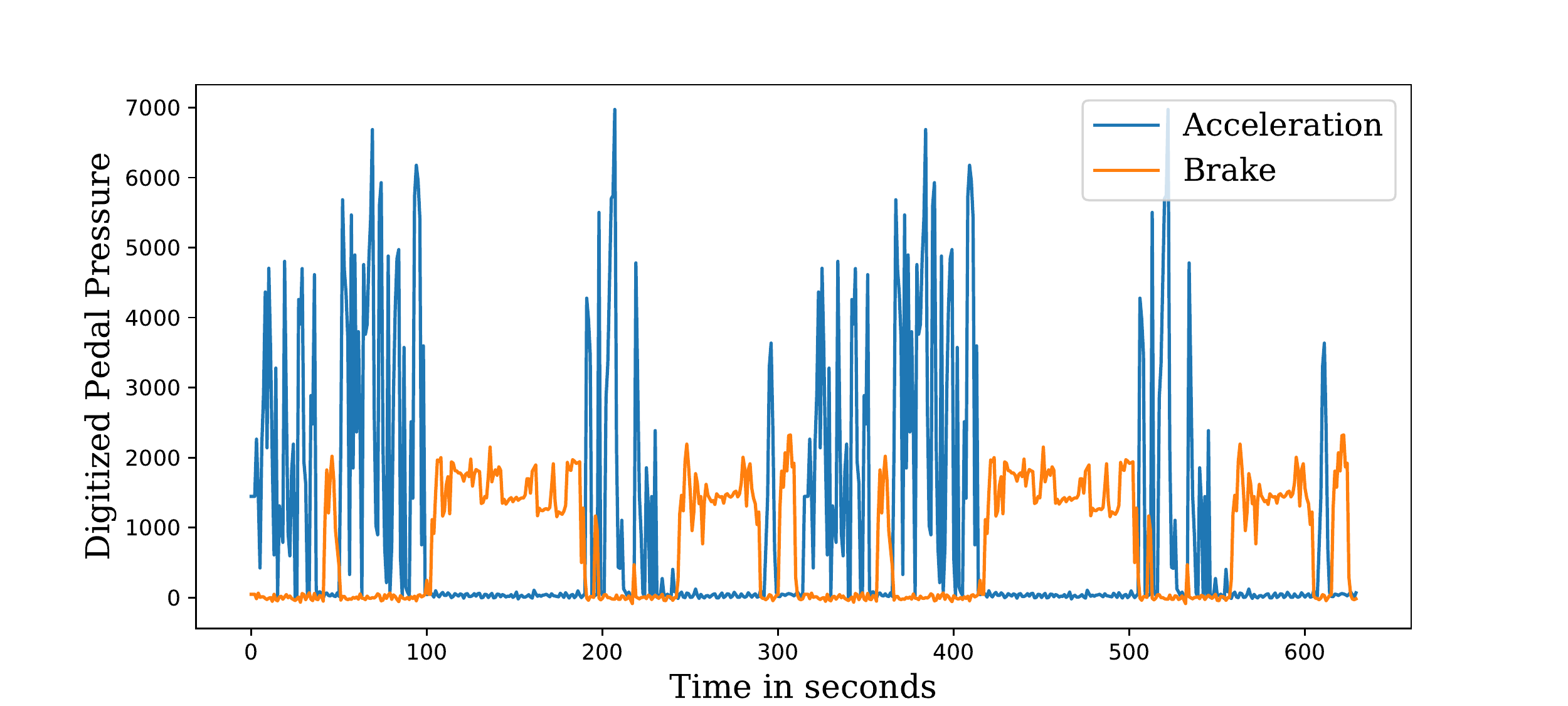} \\[\abovecaptionskip]
		(b) Driver 2
	\end{tabular}
	
	\caption{Accerleration (blue) and brake pedal pressure (orange) over time for two drivers taking the same route.}\label{fig:myfig}
	\vspace{-8.00mm} 
\end{figure}

To give the reader an idea on how this data looks like, we visualized acceleration and brake pedal pressure of two drivers from the NUDrive dataset in \Cref{fig:myfig}. Both drivers took the same route with the same car. The differences are clearly visible in this case.

According to the work of Benli et al. (2008)  and Erdogan et al.(2007), a combination of brake pedal pressure and gas pedal pressure is sufficient for driver identification and results in robust models \cite{benli2008driver, erdogan2007experiments}. Based on their experiments we only focused on pedal pressure signals.

\subsection{Evaluation Metrics}
We report two different evaluation metrics: Accuracy as well as mean per class error (\textit{MPCE}). Each driver is associated to one class. Accuracy is given by \(\frac{c}{t}\) where \(c\) is the number of correctly predicted samples and \(t\) the number of all samples.  

\textit{MPCE} is defined as the average per class error as shown in \Cref{eq:MPCE}. The parameters are the model \(m_i\), the number of class samples \(c_k\) per class \(k\) and the respective error rate \(e_k\).  

\begin{equation} \label{eq:MPCE}
\begin{array}{l}
PCE_k = \frac{e_k}{c_k} \\
\\
MPCE_i = \frac{1}{K} \sum PCE_k
\end{array}
\end{equation}

\subsection{Experiment Settings}
For the UTDrive subset we divided the data into a training, validation and test set and tuned the hyperparameters batchsize, sequence length and learning rate. For the LSTM we also tuned the size of the hidden dimension and the number of layers. Since the NUDrive subset contains very little data, we trained until convergence on a training set and evaluated on a small test set afterwards. We did not tune hyperparameters on NUDrive to save the need for a validation set. For NUDrive, our training set contains 155 sequences and for UTDrive 386 sequences.  This results in an average training time of 310 and 1554 seconds per driver respectively. Our UTDrive validation set contains 129 sequences, test sets contain 39 and 129 sequences respectively. A sequence is 20 seconds long without overlap between two sequences on which we train all networks.

To attain the same input size with equivalent sampling rate (1000 Hz vs. 100 Hz) for all the data, we use two different approaches. In our first method, we take one data point of the brake and gas signal every 0.25 seconds since the time series are almost constant in these intervals. This technique leads to information loss since all other samples are ignored. We refer to this as standard sampling technique in the following. On the other hand, Automated-Data-Feature Extraction (ADFE) uses a basic CNN to extract features. The network is automatically optimized during training and thus learns to extract the most important features from all samples.

We use cross entropy loss which combines \(log(softmax(x))\) and negative log-likelihood loss and is calculated as followed:  \(L(x, c) = -x_c + log(\sum _j e^{x_j})\), where \(x\) is the class score. The losses are averaged across observations for each minibatch.

For all experiments initial learning rate is 0.001. Learning rate is scheduled by \textit{ADAM} \cite{kingma2014adam} and batch size is 32. Furthermore, we decay the learning rate every 15 epochs by multiplying 0.5. Each model is trained for 100 epochs.

\section{Results} \label{results}
The LSTM itself reaches an accuracy of 25.64 \% and a MPCE of 75.50 \% on the NUDrive data and is therefore the model which reaches the lowest accuracy on this data. The low performance is a result of the short sequence time which makes it difficult for the LSTM to recognize long-term correlations. ResNet achieves an accuracy of 43.50 \% and a MPCE of 56.80 \%. By combining both networks as shown in \Cref{fig:LSTM-RESNET}, we obtain an accuracy of 64.10 \% and a MPCE of 32.02 \%. Furthermore, the use of ADFE instead of the standard sampling technique improves accuracy by 15.39 \% and MPCE decreases from 33.20 \% to 17.40 \%. The exact values are shown in \Cref{tab:acmpce}. 

In \Cref{tab:pcePerDriver} the amount of available training data for each driver in comparison to the respective per-class test error is visualized. It is clearly visible that the per-class error and the amount of training data are correlated. Therefore, adding more training data will increase the accuracy of our models.

\begin{table}

\caption{Accuracy and mean per-class error in \%. Indice 0 classifiers were trained with the standard sampling technique, indice 1 classifiers used ADFE.}
\begin{center}
\begin{tabular}{|c|c|c|c|}

\hline
Classifier & Dataset & AC & MPCE \\
\hline
LSTM & NUDrive & 25.64 \% & 75.50 \% \\
ResNet & NUDrive & 43.59 \% & 56.80 \% \\
LSTM-ResNet$_{0}$ & NUDrive & 64.10 \% & 32.20 \% \\
LSTM-ResNet$_{1}$ & \textbf{NUDrive} & \textbf{79.49 \%} & \textbf{17.40 \%} \\
LSTM-ResNet$_{0}$ & \textbf{UTDrive} & \textbf{96.90 \%} & \textbf{3.20 \%} \\
\hline

\end{tabular}
\end{center}
\vspace{-8.00mm} 
\label{tab:acmpce}
\end{table}

It is important that driver identification systems are easily adaptable to other drivers or cars. In order to see that our model is highly adaptable to parameter changes, we also evaluated on the UTDrive subset which contains a different amount of drivers using a different car for the data collection.

\begin{table}

\caption{Per-class error in \% for each driver and the respective amount of available training data in seconds. Results are from LSTM-ResNet$_{1}$ trained on NUDrive.}
\begin{center}
\begin{tabular}{|c|c|c|c|c|c|}

\hline
Driver ID & 1 & 2 & 3 & 4 & 5  \\
\hline
PCE & \ 0.00 \% \ & \ 67.00 \% \ & \  0.00 \% \ & \  0.00 \% \ & \  0.00 \% \ \\
Training Data & 280 & 200 & 320 & 400 & 420 \\
\hline
Driver ID & 6 & 7 & 8 & 9 & 10 \\
\hline
PCE  & 0.00 \% & 25.00 \% & 25.00 \% & 0.00 \% & 57.00 \% \\
Training Data & 260 & 260 & 260 &  400 & 300 \\
\hline
\end{tabular}
\end{center}
\label{tab:pcePerDriver}
\vspace{-8.00mm} 
\end{table}

The results on the UTDrive subset show that increasing training set size lowers test error by a large margin. The PCE in \% and the available training data per class are shown in \Cref{tab:pcePerDriverUTDrive}. We reach an identification accuracy of 96.90 \% and a MPCE of 3.20 \% with the LSTM-ResNet$_{0}$ classifier. The exact values are shown in \Cref{tab:acmpce}. These results show that the LSTM-ResNet classifier boosts the accuracy and therefore outperforms both, ResNet and LSTM on their own. The results also confirm that a basic CNN is well suited to extract features from time series.

\begin{table}

\caption{Per-class error in \% for each driver and the respective amount of available training data in seconds. Result are from LSTM-ResNet$_{0}$ trained on UTDrive.}
\begin{center}
\begin{tabular}{|c|c|c|c|c|c|}

\hline
Driver ID & 1 & 2 & 3 & 4 & 5  \\
\hline
PCE & \ 0.00 \% \ & \ 0.00 \% \ & \ 16.00 \% \ & \ 0.00 \% \ & \ 0.00 \% \  \\

Training Data & 960 & 1220 & 1580 & 1960 & 2000 \\

\hline

\end{tabular}
\end{center}
\label{tab:pcePerDriverUTDrive}
\vspace{-8.00mm} 
\end{table}

Our results outperform the results from Martinez et al. (2015) \cite{martinez2015driving} by more than 10 \% identification accuracy on a 5-drivers group. On a 11-drivers group they achieve an accuracy of 76 \%. We reach a slightly better accuracy of 79.49 \% with 10 drivers. Nishiwaki et al. (2007) reach an accuracy of 76.8 \% with a dataset containing 256 drivers \cite{nishiwaki2007driver}. Their results are difficult to compare to ours since they use a different amount of drivers, but note that their sequence length is 3 minutes compared to 20 seconds we worked with. Also note that they made use of cepstral analysis and we worked with raw pedal data.

\section{Conclusion \& Future Work} \label{sec:conclusion}
In this paper, we developed an TSC approach for identity recognition in a car cabin. By combining ResNet and LSTM in one network we achieve state-of-the-art performance with deep classifiers and reach an identification accuracy of 96.90 \% on a 5-drivers subset and 79.49 \% on a 10-drivers dataset. A big advantage in contrast to existing GMM approaches is that our method only requires to train one model for all drivers instead of one or multiple GMMs for every driver. This makes our approach suitable for an in-car environment thanks to the little memory and computing requirements. 

In future work, we would like to test our approach with other datasets in order to be able to compare to similar work. Matching dataset are the CIAIR database as well as the full UTDrive and NUDrive datasets, which were not available to us. Also, we would like to check if data collected during the first seconds after accelerating from stand is enough to classify drivers correctly. Since such data is not available in sufficient amounts it needs to be collected. Drives on highways or at same speeds vary less and it is unlikely that drivers change during driving. Therefore, such samples can be neglected. Furthermore, we would like to show that our model is easily adaptable to different amounts of drivers, even without extensive training. Transfer learning is one way to deal with this problem by using a pretrained model which can be adjusted to new drivers.






\bibliographystyle{splncs04}
%
\nocite{*}
\bibliography{bib}

\end{document}